\documentclass[10pt,twocolumn,letterpaper]{article}

\usepackage[pagenumbers]{cvpr} % To force page numbers, e.g. for an arXiv version

\usepackage{graphicx}
\usepackage{times}
\usepackage{helvet}
\usepackage{courier}
\usepackage{amsmath}
\usepackage{algorithm}
\usepackage{algorithmic}
\usepackage{csquotes} 
\usepackage{color}
\usepackage{paralist}
\usepackage{amssymb}
\usepackage{indentfirst}
\usepackage{float}
\usepackage{multirow}
\usepackage{cite}
\usepackage{mathrsfs}

\usepackage{mathrsfs}
\usepackage{textcomp,booktabs}
\usepackage{amssymb}% http://ctan.org/pkg/amssymb
\usepackage{pifont}% http://ctan.org/pkg/pifont
\usepackage[misc]{ifsym}
\newcommand{\cmark}{\ding{51}}%

\usepackage{mathrsfs}
\usepackage{color}
\usepackage{xcolor}
\definecolor{citecolor}{HTML}{0071bc} 
\definecolor{SeaGreen4}{RGB}{0,205,102} 
\definecolor{SlateBlue}{RGB}{106,90,205} 
\definecolor{DarkRed}{RGB}{178,34,34} 
\usepackage[colorlinks, linkcolor=red,  anchorcolor=blue, citecolor=citecolor]{hyperref}

\usepackage{colortbl}
\definecolor{mygray}{gray}{.9}
\definecolor{mypink}{rgb}{.99,.91,.95}
\definecolor{mycyan}{cmyk}{.3,0,0,0}

\usepackage{color}
\usepackage{xcolor}
\definecolor{citecolor}{HTML}{0071bc} 
\definecolor{SeaGreen4}{RGB}{0,205,102} 
\definecolor{SlateBlue}{RGB}{106,90,205} 
\definecolor{DarkRed}{RGB}{178,34,34}

\usepackage[capitalize]{cleveref}
\crefname{section}{Sec.}{Secs.}
\Crefname{section}{Section}{Sections}
\Crefname{table}{Table}{Tables}
\crefname{table}{Tab.}{Tabs.}

\definecolor{cvprblue}{rgb}{0.21,0.49,0.74}
\def\paperID{9748}      % *** Enter the Paper ID here
\def\confName{CVPR}
\def\confYear{2024}

\title{ MambaEVT: Event Stream based Visual Object Tracking using State Space Model }  

\author{Xiao Wang$^{1}$, Chao Wang$^{1}$, Shiao Wang$^{1}$, Xixi Wang$^{1}$, Zhicheng Zhao$^{2}$, Lin Zhu$^{3}$, Bo Jiang$^{1}$ \thanks{\Letter~~Corresponding Author: Bo Jiang} \\ 
${^1}${School of Computer Science and Technology, Anhui University, Hefei, China} \\
${^2}${School of Artificial Intelligence, Anhui University, Hefei, China} \\
${^3}${Beijing Institute of Technology, Beijing, China} \\ 
\textit{\{w853023886, wsa1943230570\}@126.com, \{wangxiaocvpr, sissiw0409\}@foxmail.com,} \\ 
\textit{zhaozhicheng@ahu.edu.cn, linzhu@bit.edu.cn, zeyiabc@163.com} \\ 
\url{https://github.com/Event-AHU/MambaEVT}
}

\begin{document}
\maketitle

%%%%%%%%% ABSTRACT
\begin{abstract}
Event camera-based visual tracking has drawn more and more attention in recent years due to the unique imaging principle and advantages of low energy consumption, high dynamic range, and dense temporal resolution. Current event-based tracking algorithms are gradually hitting their performance bottlenecks, due to the utilization of vision Transformer and the static template for target object localization. In this paper, we propose a novel Mamba-based visual tracking framework that adopts the state space model with linear complexity as a backbone network. The search regions and target template are fed into the vision Mamba network for simultaneous feature extraction and interaction. The output tokens of search regions will be fed into the tracking head for target localization. More importantly, we consider introducing a dynamic template update strategy into the tracking framework using the Memory Mamba network. By considering the diversity of samples in the target template library and making appropriate adjustments to the template memory module, a more effective dynamic template can be integrated. The effective combination of dynamic and static templates allows our Mamba-based tracking algorithm to achieve a good balance between accuracy and computational cost on multiple large-scale datasets, including EventVOT, VisEvent, and FE240hz. Source code of this work will be released. 
\end{abstract}

\section{Introduction} 

Event camera-based Visual Object Tracking (VOT) has drawn more and more attention in recent years with the release of large-scale benchmark datasets, e.g., EventVOT~\cite{wang2024event}, VisEvent~\cite{wang2023visevent}, etc. Echoing the approach of RGB-based tracking, this task aims to capture the position (\textit{x, y, w, h}) of the initialized target object within subsequent event streams as the object moves. Due to the unique imaging principle of the event camera, which outputs asynchronous streams of pulses, meaning that each pixel generates information independently without interfering with one another. It performs better than the RGB cameras on the low energy consumption, high dynamic range, and dense temporal resolution. Therefore, tracking algorithms based on event cameras are expected to be applied in more challenging scenarios, such as large-scale intelligent surveillance, military fields, and aerospace.

\begin{figure}
    \centering
    \includegraphics[width=\linewidth]{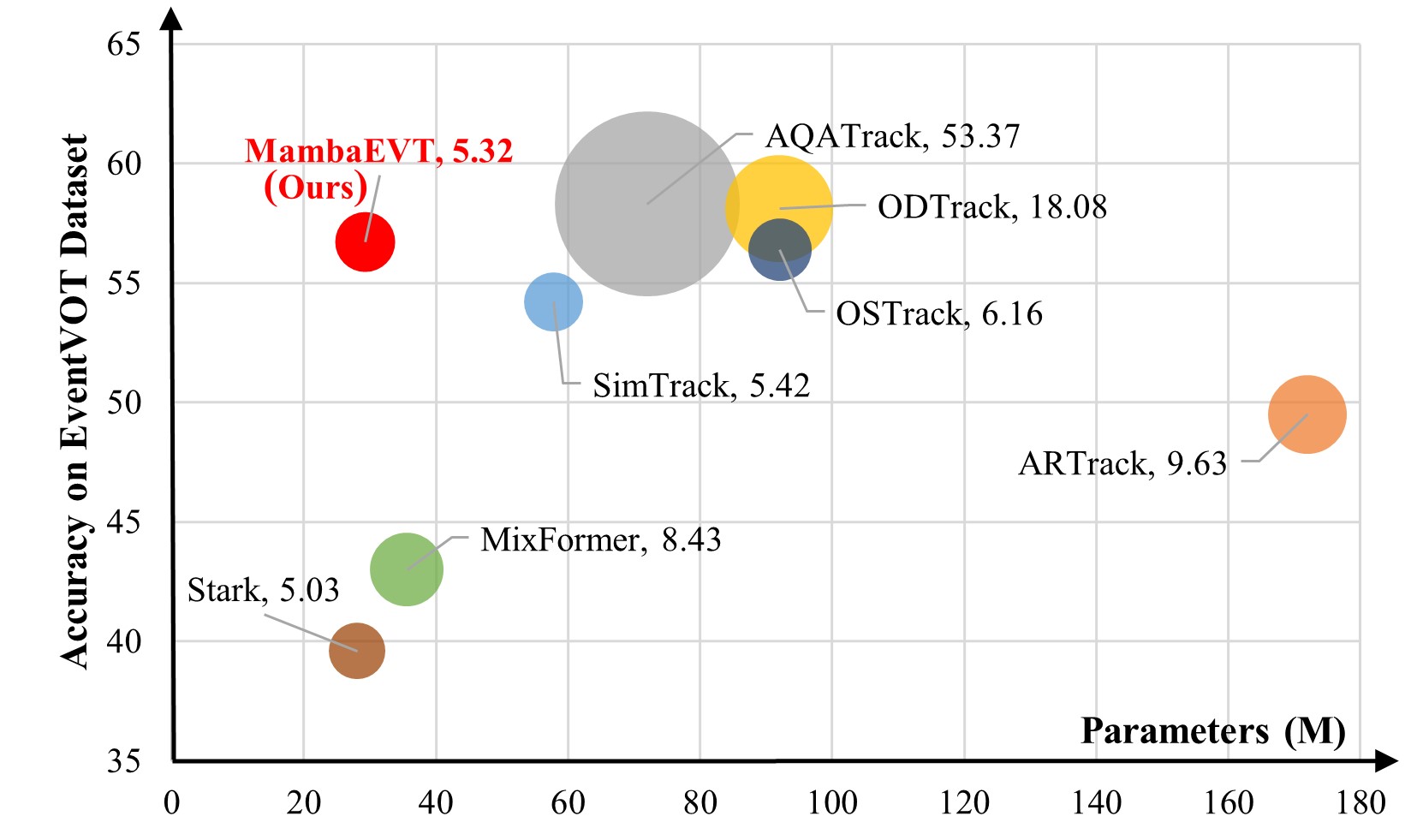}
    \caption{ Comparison between our tracker and existing SOTA trackers on the \textit{Parameters}, \textit{Accuracy}, and \textit{GPU Memory Consumption (GB)} on the EventVOT dataset. Note that, the size of the bubbles represents the amount of GPU memory usage; larger bubbles indicate a higher consumption.} 
    \label{fig:firstIMG}
\end{figure}

Tracking algorithms based on event cameras can be divided into two major categories: RGB-Event multi-modal visual tracking~\cite{hou2024sdstrack, wang2023visevent, zhang2024uniEvTrack}, and visual tracking using only the Event camera~\cite{wang2024event}. Specifically, Wang et al.~\cite{wang2023visevent} propose to fuse the RGB frames and Event streams for robust tracking using a cross-modality Transformer network. Hou et al.~\cite{hou2024sdstrack} propose the SDSTrack by adapting features from RGB modality to event-based neural networks using efficient fine-tuning ways. Zhang et al.~\cite{zhang2024uniEvTrack} propose to augment the event feature learning by boosting interactions and distinguishing alterations between states at different moments and fuse the RGB data for better tracking. HDETrack~\cite{wang2024event} proposed by Wang et al. conducts hierarchical knowledge distill from RGB-Event to Event-based tracker. 
%%%% 
To better focus on the essence of Event-based tracking and reduce interference from other factors, this paper will concentrate on the visual tracking task that utilizes only the Event camera.

By reviewing existing event-based visual trackers, we can find that these tracking algorithms are still limited by the following issues: 
\textit{1). High Computational Complexity:} Current trackers based on event streams adopt the vision Transformer (e.g., ViT~\cite{dosovitskiy2020image}) as their backbone network which is computationally expensive as validated in many works. Because the computational complexity of the self-attention mechanism in Transformer is $\mathcal{O}(N^2)$, which is highly unfriendly for the deployment of tracking algorithms on actual hardware. 
\textit{2). Static Target Template:} Existing event-based trackers follow the regular Siamese tracking framework which adopts the static template for the activation map generation. Although it works well in simple scenarios, however, the performance in long-term tracking or facing target objects with significant appearance variations will remain unsatisfactory. 
Therefore, it is natural to raise the following question: ``\textit{Can we design a new event-based visual tracker that achieves a better trade-off between tracking accuracy and computational cost?}"

Recently, the State Space Model (SSM) has become one of the most popular models in the fields of computer vision and artificial intelligence, due to its lower computational complexity $\mathcal{O}(N)$ and still commendable performance~\cite{wang2024SSMSurvey, zhu2024vision, liu2024vmamba, li2024videomamba, huang2024mambaFETrack}. 
To be specific, Vim~\cite{zhu2024vision} and VMamba~\cite{liu2024vmamba} are pioneering efforts in adapting the Mamba model for the visual domain, demonstrating its efficacy across a range of tasks. It also performs well for video-based recognition tasks, such as VideoMamba~\cite{li2024videomamba}. 
% For the tracking task, Huang et al.~\cite{huang2024mambaFETrack} develop a new RGB-Event tracking framework based on the Mamba network, termed Mamba-FETrack, and achieve good results on multiple RGB-Event tracking benchmark datasets. 
Inspired by these works, in this paper, we propose a novel Mamba-based Visual Object Tracking algorithm based on Event cameras, termed \textit{MambaEVT}. As shown in Fig.~\ref{fig:framework}, we first extract the static template and search regions from the given event streams and project them into event tokens. Then, the vision Mamba is adopted for feature extraction and interactive learning. In this procedure, both forward and backward directions are taken into account, allowing the learned visual features to be free from directional interference, resulting in more accurate and robust. The output event tokens will be fed into a tracking head for target object localization. Similar operations are conducted until the end of the event stream.

More importantly, we design a new adaptive update strategy for \textit{dynamic template generation} using a Memory Mamba. In our implementation, we collect the features corresponding to tracking results and maintain a template library, deciding whether to update it based on appearance diversity. Leveraging the powerful modeling capabilities of the Mamba model for long sequences, we propose Memory Mamba to fuse the sample features within the target template library, thereby accomplishing the generation of dynamic templates. This assists the tracker in better adapting to significant appearance changes of the object, minimizing their impact on the tracking results. Unlike conventional dynamic update mechanisms~\cite{yan2021stark} (which are only deployed during the test phase and are not learnable), our method is trainable, allowing for better utilization of training data to enhance its modeling capabilities. Consequently, during the practical tracking phase, we perform feature learning on both the dynamically updated templates and the initially given static templates to achieve more accurate and robust tracking results.

To sum up, the main contributions of this paper can be summarized as the following three perspectives: 

1). We propose a novel Mamba-based event tracking framework, termed \textit{MambaEVT}, which achieves a good trade-off between accuracy and computational cost. To the best of our knowledge, it is the first Mamba-based tracking framework using an Event camera. 

2). We propose a simple but effective dynamic template update strategy using the Memory Mamba network. Our template update module is learnable and further enhances the final tracking performance. 

3). Extensive experiments on multiple large-scale event-based tracking datasets (i.e., EventVOT, VisEvent, FE240hz) fully validated the effectiveness and efficiency of our proposed Mamba-based visual object tracking framework.

\section{Related Work}

\subsection{Event based Visual Object Tracking} 
Recently, event-based visual object tracking has emerged as a significant research topic. In early works, Huang et al.~\cite{huang2018event} proposed an event-guided support vector machine (ESVM) for tracking high-speed moving objects. 
Glover et al.~\cite{glover2017robust} introduced a particle filter designed for event-based tracking in dynamic environments using event cameras. 
Mitrokhin et al.~\cite{mitrokhin2018event} introduced an object-tracking method using event-based vision sensors that capture dynamic data and detect moving objects. 
Gehrig et al.~\cite{gehrig2018asynchronous} presented a method that combines event cameras with standard cameras for enhanced visual tracking using a maximum-likelihood approach, resulting in more accurate and longer tracks. 
Jiang et al.~\cite{jiang2021flow} introduced a new ``motion feature" derived from event cameras, utilizing the Surface of Active Events to capture pixel-level motion on the image plane. 
The work of Jiao et al.~\cite{jiao2021comparing} compare two types of image representations for event camera-based tracking and introduce a check for weaknesses, along with an improved tracker that combines the strengths of both representations. 
Zhang et al.~\cite{zhang2021object} introduce a novel frame-event fusion approach for single object tracking, enhancing performance in challenging conditions such as high dynamic range, low light, and fast motion. 
Then, Zhang et al.~\cite{zhang2022spiking} proposed a spiking Transformer network (STNet) for single object tracking, effectively extracting and fusing temporal and spatial information. 
KLT Tracker~\cite{messikommer2023data} introduced the first data-driven feature tracker for event cameras, integrating low-latency events with grayscale frames. 
Wang et al.~\cite{wang2023visevent} propose a simple and effective baseline tracker, which can fully utilize the unique information of different modalities for robust tracking by developing a cross-modal transformation module. 
Tang et al.~\cite{tang2022revisiting} propose CEUTrack, which is an adaptive unified tracking method based on the Transformer network with a one-stage backbone network. 
Wang et al.~\cite{wang2024event} propose a novel hierarchical cross-modality knowledge distillation approach for an event-based tracking problem, called HDETrack. They exploit the knowledge transfer from multi-modal / multi-view to an unimodal event-based tracker. 
%%%% 
Unlike the mainstream Transformer-based tracking algorithms prevalent in recent years, our work stands out by exclusively adopting a pure Mamba-style approach for efficient tracking. Our Memory Mamba network also helps the event-based tracking via dynamic template update.

\subsection{Dynamic Template Updating}  
In the field of visual object tracking, the dynamic updating of template information is frequently adopted according to the mechanisms of memory networks. In the early stages of research, 
Zheng et al.~\cite{6305743} enhanced video tracking systems by combining dynamic template updating with Kalman filter-based tracking. 
Yang et al.~\cite{yang2019visual} proposed a dynamic memory network that uses an LSTM with spatial attention to adapt the template to target appearance changes during tracking. 
Memformer~\cite{wu2020memformer} is an efficient neural network for sequence modeling, which utilizes a fixed-size external dynamic memory to encode and retrieve past information. It efficiently processes long sequences with linear time complexity and constant memory complexity. 
STARK~\cite{yan2021stark} updates the dynamic template to capture appearance changes of tracking targets, rather than updating object queries like typical Transformer-based models. 
Liu et al.~\cite{liu2021effective} propose a template update mechanism designed to enhance visual tracking accuracy, specifically to address the issue of tracking failure in cluttered backgrounds. 
Cui et al.~\cite{cui2022Mixformer} propose a Score Prediction Module (SPM) to select reliable online templates based on predicted scores, achieving efficient online tracking. 
Yang et al.~\cite{10113336} propose a Siamese tracking network with dynamic template updating to adapt to appearance changes, improving accuracy in tracking small objects in satellite videos. 
Zhang et al.~\cite{zhang2024robust} propose a robust tracking method centered on dynamic template updating (DTU), enhancing the original Siamese network with new branches and feedback links. 
Wang et al.~\cite{wang2021DeepMTA} also introduce tracking results to build a dynamic template pool for accurate object tracking. 
%%%%
Different from previous studies, we incorporate a Memory Mamba network to serve as the dynamic template updating module, designed to capture the temporal changes in the appearance features of targets. Our goal is to improve tracking performance with a simple and efficient method. 
%aimed at capturing the temporal variations of the appearance features of targets. We endeavor to enhance tracking performance through a straightforward and efficient approach. 

\subsection{State Space Model} 
Research in state space models (SSMs) has also surged recently. It inherits the ability of the classic Structured State-Space Sequence (S4)~\cite{gu2021efficiently} model to handle long sequences and enhances content-based reasoning. However, the initial SSMs could only be used for one-dimensional sequence modeling, such as text sequence modeling in the field of NLP. Subsequently, Nguyen et al.~\cite{nguyen2022s4nd} propose to extend 1D sequence modeling to 2D and 3D visual tasks such as images and videos. Furthermore, Gu et al.~\cite{gu2023mamba} introduce a selective scanning mechanism based on SSM, called Mamba, which greatly improves the modeling ability of sequences. Based on this, a large number of visual works based on Mamba have flooded in. 
Vision Mamba~\cite{zhu2024vision} and VMamba~\cite{liu2024vmamba} have successively demonstrated the effectiveness of Mamba in visual tasks, which using a bidirectional scanning mechanism and a four-way scanning mechanism, respectively. Subsequently, in the field of video, Video Mamba~\cite{li2024videomamba} added the dimension of time to handle 3D video tasks and effectively model video sequences. 
In this work, we propose a fully Mamba-style framework based on Vision Mamba. In addition to the Mamba module used for feature extraction, an additional Mamba module generates high-quality dynamic template information through a dynamic template library, which then serves as essential support for subsequent feature extraction. The design enables efficient parameter utilization and superior performance in event-based tracking. 
%which then serves as critical auxiliary information for subsequent feature extraction. 

\section{Methodology}  
\begin{figure*}
\centering
\includegraphics[width=1\linewidth]{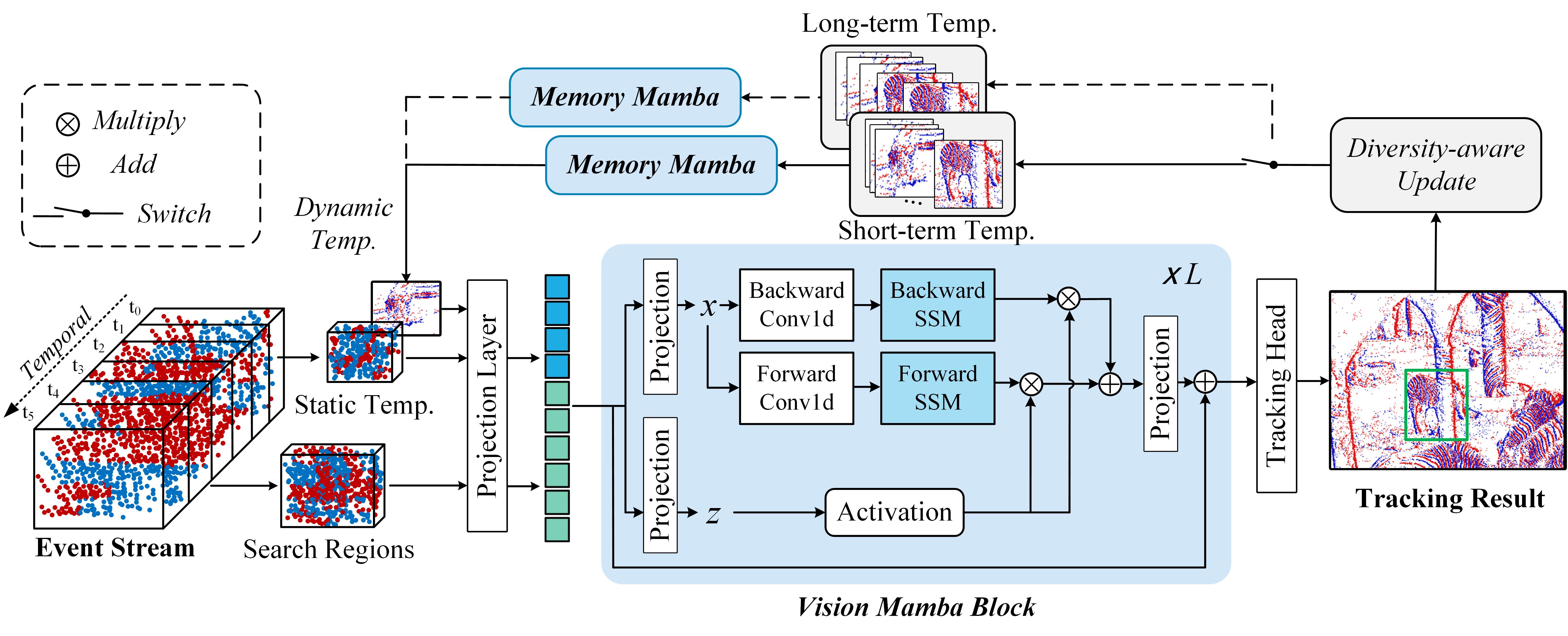}
\caption{
An overview of our proposed pure Mamba-based visual object tracking using an event camera, termed MambaEVT. The first key feature of our tracking framework is the vision Mamba based backbone network which can achieve feature extraction, interaction and fusion, simultaneously. It ensures our tracker achieves good performance and also lowers computational cost. The second one is the learnable Memory Mamba for dynamic template generation which makes our tracker more robust to significant appearance variation. 
} 
\label{fig:framework}
\end{figure*}

\subsection{Preliminary: Mamba}   

The raw state space model is developed for the continuous system. It maps a one-dimensional function or sequence \( x(t) \in \mathbb{R} \) to \( y(t) \in \mathbb{R} \) through a hidden state \( h(t) \in \mathbb{R}^N \). In this system, the evolution of the hidden state is governed by the matrix \( \mathbf{A} \in \mathbb{R}^{N \times N} \) and the input is projected by the matrix \( \mathbf{B} \in \mathbb{R}^{N \times L} \), while the output projection is given by the matrix \( \mathbf{C} \in \mathbb{R}^{L \times N} \). The formula is as follows,
\begin{align}
    h'(t) &= \mathbf{A} h(t) + \mathbf{B} x(t), \\
    y(t) &= \mathbf{C} h(t).
\end{align}
To facilitate the understanding of computer operating systems, the discrete counterparts of S4~\cite{gu2021efficiently} allow the transformation of the continuous parameters \( \mathbf{A} \) and \( \mathbf{B} \) into discrete parameters \( \bar{\mathbf{A}} \) and \( \bar{\mathbf{B}} \). This transformation is typically performed using the Zero-Order Hold (ZOH) method:
\begin{align}
    \bar{\mathbf{A}} &= \exp(\Delta \mathbf{A}), \\
    \bar{\mathbf{B}} &= (\Delta \mathbf{A})^{-1} (\exp(\Delta \mathbf{A}) - \mathbf{I}) \cdot \Delta \mathbf{B}, 
\end{align}
where \( \Delta \) is a discrete time step. Therefore, the discrete system can be reformulated as,
\begin{align}
    h_t &= \bar{\mathbf{A}} h_{t-1} + \bar{\mathbf{B}} x_t, \\
    y_t &= \mathbf{C} h_t.
\end{align}

As we know, the context information is stored in the similarity matrix in the Transformer. However, the Structured State Machine (SSM) lacks a comparable module, which leads to sub-optimal performance in contextual learning tasks. To address this issue, Gu et al.~\cite{gu2023mamba} introduce the Mamba architecture, which enhances the SSM in two significant ways, i.e., \textit{Selective Scan Operator}, and \textit{Hardware-aware Algorithm}. Thanks to the Mamba network's powerful modeling ability for visual information, in this paper, we build a pure Mamba-based single object tracking algorithm that achieves a better trade-off between accuracy and computational cost.

% \begin{itemize}
%     \item : This operator selectively filters relevant information by making $\Delta$, $\textbf{B}$, and $\textbf{C}$ functions of the input while keeping $\textbf{A}$ unchanged.
%     \item : The algorithm efficiently stores intermediate results through techniques such as parallel scanning, kernel fusion, and recalculation.
% \end{itemize}

\subsection{Overview}  
As shown in Fig.~\ref{fig:framework}, our proposed event-based visual tracker, i.e., MambaEVT, follows the Siamese tracking framework. We stack the event streams into event frames and crop the target template and search region out as the input. Then, a projection layer is adopted to embed these inputs into event tokens and the output will be fed into the vision Mamba backbone network for feature extraction and interactive learning. We propose a tracking head that takes the tokens corresponding to the search region as the input for target object localization. In addition, we also propose a new dynamic template generation module using Memory Mamba to address the significant appearance variations in the tracking task. More details of the modules mentioned in this procedure will be introduced in the subsequent sub-sections, respectively.

\subsection{Input Representation}  
In this work, we denote the event streams as $\mathcal{E} =\left\{e_{1} ,e_{2},..., e_{M}\right\}$, where $e_{i} = [x, y, t, p]$ represents each event point asynchronously launched, $(x, y)$ denotes the spatial coordinates, $t$ is the timestamp, $p$ is the polarity (i.e, ON/OFF event), $i \in $$\displaystyle [ 1, \ M]$, $\displaystyle \mathnormal{M}$ is the number of event points in the current sample. We obtain the event image $\mathcal{E}_F \in \mathbb{R}^{3 \times  W \times H }$ by splitting the event stream into fixed-length sequences and using these event images as the inputs to the framework.

As mentioned above, our tracker follows the Siamese framework which takes the template region and search region as the input. Thus, the initial template extracted from the first frame is denoted as  $\mathcal{E}_{z}^{0} \in \mathbb{R}^{3 \times W_{z} \times H_{z}}$,  and the search region is denoted as $\mathcal{E}_{x}^{t} \in \mathbb{R}^{3\times W_{x} \times H_{x}}$. We also maintain a dynamic template using the Memory Mamba network to handle the appearance variation issue which will be introduced later and represent it as $\mathcal{E}_{d}^{t} \in \mathbb{R}^{N\times C}$, where $\displaystyle N=\frac{W_{z} \times H_{z}}{P^{2}}$, $P$ is the resolution of each patch and $\displaystyle \mathnormal{C}$ is the dimension of the token.

\subsection{Mamba-based Tracking}  
Given the search region $\mathcal{E}_{x}^{t}$ and static template $\mathcal{E}_{z}^{0}$, we first embed and flatten them into one-dimensional tokens, denoted as $S_{x}^{t} \in \mathbb{R}^{N_{x} \times C}$ and $T_{z}^{0} \in \mathbb{R}^{N_{z} \times C}$, respectively. Here, $\displaystyle N_{z} =\frac{W_{z} \times H_{z}}{P^{2}}$ and $\displaystyle N_{x} =\frac{W_{x} \times H_{x}}{P^{2}}$. Note that, the position embeddings are then added to the patch embeddings of both the initial template and the search region. The resulting token sequences are concatenated with dynamic features to form $\mathcal{E}_{f} =[T_{z}^{0}, \mathcal{E}_{d}^{t}, S_{x}^{t} ]$. Inspired by the success of unified backbone networks for tracking, e.g., OSTrack~\cite{ye2022joint} and CEUTrack~\cite{tang2022revisiting}, we pass the concatenated feature $\mathcal{E}_{f}$ through the Vision Mamba block (\textit{Vim}, for short)~\cite{zhu2024vision} backbone to achieve feature extraction, interaction, and fusion, simultaneously. 
Let's denote the input of the $l$-th \textit{Vim} layer is $\mathcal{E}_{f}^{l-1}$, the forward propagation procedure can be formulated as: 
\begin{equation}
\label{eq:eq1}
\mathcal{E}_{f}^{l} = Vim \left( \mathcal{E}_{f}^{l-1} \right) + \mathcal{E}_{f}^{l-1}, ~~l=\{1, 2, \ldots, L\}. 
\end{equation}
Then, we normalize the output tokens $\mathcal{E}_{f}^{l}$ and feed them into the multi-layer perception (MLP) to produce the final features, i.e., 
\begin{align}
\bar{\mathcal{E}}_{f}^{l} = MLP(Norm(\mathcal{E}_{f}^{l})), 
\end{align}
We adopt both forward and backward directions to learn a more robust deep representation to make our tracker free from directional interference, as illustrated in Fig.~\ref{fig:framework}.

Finally, we crop the event tokens corresponding to search regions from the enhanced feature $\mathcal{E}_{f}$, denoted as $S_{x}^{t}$, and feed them into the tracking head. In our implementation, the tracking head is implemented as a Fully Convolutional Network (FCN), which is composed of a series of stacked \texttt{Conv}-\texttt{BN}-\texttt{ReLU} layers. 
The FCN produces three key outputs: 
1) a target classification score map that indicates the likelihood of each spatial location being the target. 
2) a local offset to correct for discretization errors due to reduced resolution. 
3) a normalized bounding box size that specifies the width and height for the final target detection.

\subsection{Dynamic Template Update Strategy} \label{DynaTempUpate} 
Although the aforementioned Mamba-based event tracker already achieves a good performance, however, it is still sensitive to challenging factors such as significant appearance variation, and clutter background. Obviously, the tracking performance can be boosted when these issues can be handled. In this work, we propose the Memory Mamba to collect the tracking results and generate an adaptive dynamic template to guide the tracker. Two core aspects need to be considered when designing this module, i.e., 
\textit{Template Library Update}, 
\textit{Dynamic Template Generation} via Memory Mamba.

\noindent $\bullet$ \textbf{Template Library Update.~} 
To achieve dynamic template generation, we need to maintain a template library (i.e., image patches corresponding to tracking results in the inference phase). Inspired by THOR~\cite{sauer2019tracking}, we also propose to store the templates using Long-Term memory (LT) and Short-Term memory (ST) libraries. Specifically, the ST is a queue-like list that takes in the cropped image from the prediction of the search image, sampling templates at fixed intervals. The LT handles long-range memory, sampling based on a similarity-based strategy. The incoming template is evaluated against the existing features stored in both memory libraries. The similarity of the new template is determined by comparing it with each feature in both the long-term and short-term memory libraries. After identifying the pair with the highest similarity, the corresponding memory library, either LT or ST, is selected and sent as a sequence into the Memory Mamba network to generate a dynamic feature. 

The new dynamic template is added to the ST at regular intervals. Once the ST memory is at full capacity, which is typically fully initialized, a template is removed. The removed template is transferred to the LT only if replacing an existing template increases the Gram determinant of the LT's current matrix. Formally, for a new template \( z_{new} \), the LT is a set of \( \{z_{1}, z_{2}, ..., z_{n}\} \), where \( n \) is the capacity of the LT. The Gram matrix of the LT can be formulated as: 
\begin{equation}
\begin{aligned}
G(z_{1}, \cdots, z_{n}) &= 
\begin{bmatrix}
z_{1} \star z_{1} & z_{1} \star z_{2} & \cdots & z_{1} \star z_{n} \\
\vdots & \vdots & \ddots & \vdots \\
z_{n} \star z_{1} & z_{n} \star z_{2} & \cdots & z_{n} \star z_{n}
\end{bmatrix}
\end{aligned}
\end{equation}
where \( G \) is a square \( n \times n \) matrix and \( \star \) denotes the similarity measure method. 
Approximately, the determinant of \( G \) can be regarded as a measure of the diversity of the matrix. 
In this work, we utilize the Pearson linear correlation method as the similarity measure for paired template features.

\noindent $\bullet$ \textbf{Dynamic Template Generation via Memory Mamba.~} 
Given the short-term and long-term memory library, we propose a new Memory Mamba network to fuse these templates into a single one considering its strong modelling ability for the long sequence. The templates in the memory library are treated as a sequence \( Z = [z_{1}, z_{2}, \ldots, z_{m}] \), where \( z_{i} \in \mathbb{R}^{N \times C} \) and \( m \) represents the number of templates in either the LT or ST memory. The sequence will be fed into the Memory Mamba network (denoted by $\texttt{Mem-Mamba}$), whose architecture is the same as the vision backbone. The forward propagation process within the memory is formulated as follows:
\begin{equation}
    Z^{l} = \texttt{Mem-Mamba}(Z^{l-1}),  ~~ l= \{1, 2, \ldots, L\}
\end{equation}
We take the output from the last layer \( Z^{l} \in \mathbb{R}^{m \times N \times C} \) and use the last \( N \) patches as the fused dynamic template $\mathcal{E}_{d}$. Since the Mamba block is highly effective at processing sequential data, we assume that the final part of the sequence contains the most relevant information. Note that, short-term and long-term memory share the same $\texttt{Mem-Mamba}$ for dynamic template generation.  

% The primary distinction between our method and THOR in generating dynamic features lies in the approach to utilizing information. Our method leverages all available information from the corresponding memory library, whereas THOR only uses the features from the pair with the highest similarity.

\subsection{Loss Function}   
The loss functions utilized in our framework consist of a weighted combination of 
\textit{focal loss} $\mathcal{L}_{focal}$, \textit{L1 loss} $\mathcal{L}_{1}$, and \textit{Generalized Intersection over Union (GIoU) loss} $\mathcal{L}_{GIoU}$. The overall loss function is defined as follows:
\begin{equation} 
\mathcal{L} = \lambda_{1} L_{1} + \lambda_{2} \mathcal{L}_{focal} + \lambda_{3} \mathcal{L}_{GIoU}
\end{equation}
where the trade-off parameters are set as $\lambda_{1} = 5$, $\lambda_{2} = 1$, and $\lambda_{3} = 2$ in our experiments. 

\section{Experiment}

\begin{table*}
\centering
\small     
\resizebox{\textwidth}{!}{
\begin{tabular}{cccccccccc}
\toprule 
\textbf{DiMP}~\cite{zhang2022spiking}  &\textbf{TransT}~\cite{chen2021transformer}  &\textbf{STARK}~~\cite{yan2021stark}    &\textbf{PrDiMP}~\cite{danelljan2020probabilistic}   &\textbf{SiamFC++}~\cite{xu2020siamfc++} &\textbf{SiamPRN}~\cite{li2018high}           \\ 
53.4/88.2       &56.7/89.0        &55.4/83.7       &54.5/85.3            &53.38/83.89        &41.6/75.5        \\ 
\midrule 
\textbf{HIPTrack}~\cite{cai2024hiptrack}  &\textbf{OSTrack}~\cite{ye2022joint}    &\textbf{ARTrack}~\cite{wei2023autoregressive}  &\textbf{AQATrack}~\cite{xie2024autoregressive}  &\textbf{MambaEVT} &\textbf{MambaEVT-P}      \\ 
55.2/86.8      &57.1/89.3          &59.21/91.40        &59.86/92.57    &58.09/91.97    &57.24/90.37\\         
\bottomrule 
\end{tabular}
}
\caption{Experimental results (SR/PR) on FE240hz dataset.} 
\label{tab:FE240table}
\end{table*}

\subsection{Datasets and Evaluation Metrics} 
To validate the effectiveness and superiority of the proposed MambaEVT method, in this work, we conduct extensive experiments on three event-based tracking datasets, including \textbf{EventVOT}~\cite{wang2024event}, \textbf{FE240hz}~\cite{zhang2021object} and \textbf{VisEvent}~\cite{wang2023visevent}. 
We provide a brief introduction to these event-based tracking datasets in our supplementary materials.

For the evaluation metrics, we use the widely recognized \textbf{Precision Rate (PR)}, \textbf{Normalized Precision Rate (NPR)}, and \textbf{Success Rate (SR)}.

\subsection{Implementation Details} 

Our proposed tracker is implemented using Python based on PyTorch framework. The experiments are conducted on a server with NVIDIA RTX 3090 GPUs. More in detail, two variants are proposed, i.e., MambaEVT and MambaEVT-P. MambaEVT indicates that the Memory Mamba ($\texttt{Mem-Mamba}$) module and the backbone network share parameters, while MambaEVT-P indicates that they do not. In our experiments, the MambaEVT is adopted as the default tracker, unless otherwise noted. During the training phase, we resize the templates and search regions to $128 \times 128$ and $256 \times 256$ pixels, respectively. We initialize the backbone network using the Vim-S model pre-trained on ImageNet-1K dataset and train it for 50 epochs. We use AdamW~\cite{loshchilov2017decoupled} optimizer with weight decay $1 \times 10^{-4}$. The initial learning rate is set to 0.0004, with a decline beginning after 30 epochs.

For the training sample preparation, we denote the number of dynamic templates as $K$ and set $K$ to 7 during the training phase by default. We randomly sampled $K+2$ images from the entire sequence, selecting $K$ as dynamic templates, and the remaining images served as the static template and search region, respectively. This design aims to stimulate different states of long-term (LT) or short-term (ST) for better interactive learning. 
During testing, both LT and ST are fully populated with the initial template from the starting frame of the tracking process and are subsequently updated at regular intervals using the Memory Mamba network. The experimental results in the subsequent sections demonstrate the effectiveness of the consistency of training and testing. The sampling interval is set to 5. Both LT and ST memories have a default capacity of 10. More details can be found in our source code.

\subsection{Comparison with the State-of-the-art}

\begin{table}
\centering
\small   
\resizebox{\columnwidth}{!}{ 
\begin{tabular}{lcccccc} 
\toprule
\textbf{Trackers} & \textbf{Source}   & \textbf{SR}  &\textbf{PR}   &\textbf{NPR}  &\textbf{Params}  &\textbf{FPS}\\ 
\midrule
\textbf{ TrDiMP}~\cite{wang2021transformer}& CVPR21     &\ 39.9   &35.3&47.2&\ 26.3   &\ 26   \\ 
\textbf{ ToMP50}~\cite{mayer2022transforming}   &  CVPR22   &\ 37.6   &33.5&45.6&\ 26.1   &\ 25  \\ 
\textbf{ OSTrack }~\cite{ye2022joint}   &  ECCV22   &\ 55.4  &56.4&65.2&\ 92.1   &\ 105  \\ 
\textbf{ STARK }~\cite{yan2021stark}   &  ICCV21     &\ 44.5   &\ 39.6  &52.0&\ 28.1   &\ 42  \\ 
\textbf{ TransT}~\cite{chen2021transformer}   &  CVPR21     &\ 54.3  &53.5&63.2&\ 18.5   &\ 50  \\ 
\textbf{ MixFormer}~\cite{cui2022Mixformer}   & CVPR22     &\ 49.9   &43.0&53.7&\ 35.6   &\ 25  \\ 
\textbf{ SimTrack}~\cite{chen2022backbone}   & ECCV22     &\ 55.4   &54.2&64.1&\ 57.8   &\ 40  \\ 
\textbf{ ARTrack}~\cite{wei2023autoregressive} & CVPR23& 52.0& 49.5& 60.3& 172&26\\ 
\textbf{ AQATrack}~\cite{xie2024autoregressive} & CVPR24  &57.8&58.3&66.6& 72&68\\ 
\textbf{ ODTrack}~\cite{zheng2024odtrack} & AAAI24& 56.8& 58.1& 66.2& 92&32\\ 
\hline 
\textbf{ MambaEVT (Ours) }&--  &56.5&56.7&65.5&29.3&28\\
\textbf{ MambaEVT-P (Ours) } &-- & 55.4& 55.9& 64.2& 54.7&25\\ 
\bottomrule
\end{tabular}
}
\caption{Overall tracking performance on EventVOT dataset. } 
\label{tab:EventVOT_auc}
\end{table}

\noindent $\bullet$ \textbf{Results on EventVOT dataset.~} 
% state-of-the-art (SOTA)
To verify the effectiveness of our proposed method, we compare several SOTA trackers on the EventVOT dataset. 
As shown in Table~\ref{tab:EventVOT_auc}, our proposed MambaEVT achieves the performance of 56.5 in SR metric, 56.7 in PR metric, and 65.5 in NPR metric, respectively. 
Compared with OSTrack~\cite{ye2022joint} method based on ViT~\cite{dosovitskiy2020image} model, our proposed MambaEVT improves the SR/PR/NPR scores by 1.1, 0.3, and 0.3 points, while significantly reducing the number of parameters from 92.1M to 29.3M. 
Although ODTrack~\cite{zheng2024odtrack} and AQATrack~\cite{xie2024autoregressive} show better performance with scores of 57.8/58.3/66.6 and 56.8/58.1/66.2, they require significantly more parameters, totaling 72M and 92M. 
Our proposed MambaEVT needs just 29.3M parameters while still achieving competitive tracking performance. 
These results demonstrate that the proposed MambaEVT model is not only effective but also parameter-efficient.

\noindent $\bullet$ \textbf{Results on FE240hz dataset.~} 
We compare our proposed MambaEVT method with several SOTA trackers on the FE240hz dataset in Table~\ref{tab:FE240table}. 
Our propose MambaEVT achieves the performance of 58.09 in SR metric and 91.97 in PR metric.
%When compared with Transformer-based methods such as AQATrack (59.86/92.57) and ARTrack (59.21/91.40), our propose MambaEVT outperforms ARTrack~\cite{wei2023autoregressive} on the PR metric and is slightly lower than AQATrack on the same metric.  
While AQATrack and ARTrack achieve higher performance, they require significantly more parameters. Our proposed MambaEVT, on the other hand, strikes an excellent balance between performance and parameter efficiency. Specifically, it outperforms ARTrack on the PR metric by 0.57 and falls short of AQATrack on the same metric by only 0.60. 
These results highlight the effectiveness of our proposed approach.

\noindent $\bullet$ \textbf{Results on VisEvent dataset.~} 
As shown in Table~\ref{tab:VisEvent}, we compare our proposed MambaEVT with several SOTA trackers on the VisEvent dataset. 
MambaEVT achieves SR/PR/NPR scores of 35.9/50.2/39.4, while its variant, MambaEVT-P, attains higher scores of 37.2/51.8/40.8. This improvement is likely due to the need for more parameters to capture key features when handling variable-length sequences on the VisEvent dataset.  
While the top trackers, HIPTrack (37.2/52.0/41.1) and AQATrack (38.7/53.6/42.6), achieve higher scores, MambaEVT shows competitive tracking performance with significantly fewer parameters. 
These results underscore the strength of our method in balancing performance with parameter efficiency, making it a scalable and efficient solution across various scenarios, despite its slightly lower performance.

\begin{table}
\centering
\small   
\resizebox{\columnwidth}{!}{ 
\begin{tabular}{lcccc}
\toprule
\textbf{Trackers} & \textbf{SR}  & \textbf{PR}  & \textbf{NPR} &\textbf{Params} \\
\midrule 
\textbf{STARK }~\cite{yan2021stark}            & 34.8 & 41.8 & - &28.1\\ 
\textbf{OSTrack }~\cite{ye2022joint}          & 34.5 & 48.9 & 38.5 &92.1\\ 
\textbf{ARTrack }~\cite{wei2023autoregressive}            &33.9 &47.0 &37.2 &172.0 \\
\textbf{ODTrack}~\cite{zheng2024odtrack} &35.4 &50.2 &39.4 &92.0\\
\textbf{HIPTrack}~\cite{cai2024hiptrack}  &37.2 &52.0 &41.1 &120.4\\
\textbf{AQATrack }~\cite{xie2024autoregressive}  &38.7 &53.6 &42.6 &72.0 \\
\hline 
\textbf{MambaEVT }          &35.9  &50.2  &39.4 &29.3\\  
\textbf{MambaEVT-P }        &37.2  &51.8  &40.8 &64.2\\ 
\toprule
\end{tabular}
}
\caption{Event-Only tracking performance on VisEvent. } 
\label{tab:VisEvent}
\end{table}

\subsection{Ablation Study}  

\begin{table}
\centering
\small     
% \resizebox{\columnwidth}{!}{
\begin{tabular}{ccccc} 
\toprule
\textbf{Variants}  & \textbf{Mamba} & \textbf{MM} & \textbf{ML}& \textbf{SR/PR} \\ 
\midrule  
1 &\cmark   &          &         &55.0/55.2\\  
2 &\cmark   &\cmark    &         &56.3/56.1\\  
3 &\cmark   &          &\cmark   &54.2/54.1\\  
4 &\cmark   &\cmark    &\cmark   &\textbf{56.5/56.7}\\  
\bottomrule 
\end{tabular}
% } 
\caption{Ablation study of different components in the proposed MambaEVT model on EventVOT dataset. `MM' denotes two Memory Mamba modules with shared parameters and `ML' stands for Memory Library.} 
\label{tab:CAResults} 
\end{table}

\noindent $\bullet$ \textbf{Component Analysis.~}
To analyze the effectiveness of different components in the proposed MambaEVT model, we implement three variants for the analysis of the ablation study. 
`Variant \#1' model indicates that we just use the Mamba model to replace the backbone network in~\cite{ye2022joint} as the baseline network. 
`Variant \#2' model represents we added a Memory Mamba (MM) module with shared parameters in the `Variant \#1' model. 
% `Variant #2' model represents that we add the Memory Mamba (MM) module, with parameters shared with the `Variant #1' model.
`Variant \#3' model integrates the Memory Library (ML) strategy into the `Variant \#1' model, facilitating the selection of long-term and short-term templates. 
`Variant \#4' model combines MM module and ML strategy within `Variant \#1'model, i.e., MambaEVT model. 
The experimental results are presented in Table~\ref{tab:CAResults} on the EventVOT dataset.
We can observe that, 1) Compared with `Variant \#1' and `Variant \#2', it demonstrates that the incorporation of the MM module significantly improves the model performance.
% 2) Compared with `Variant \#1' and `Variant \#3' model, the results indicate that the application of the ML strategy improves the model performance.
2) Compared with the `Variant \#1' and `Variant \#3' model, the results indicate that the application of the Memory Library strategy led to a decline in model performance, likely due to a misalignment between the training and testing phases. 
3) When both the Memory Mamba module and the Memory Library strategy are integrated into the baseline network (i.e., `Variant \#4'), the performance surpasses that of the other three variants. 
In other words, these results demonstrate the effectiveness of each component and that they are used together to achieve the best model performance.

\begin{table}
\centering
\small 
\begin{tabular}{c|ccccc}
\toprule
& \textbf{ViT-S}& \textbf{Swin-S} & \textbf{VM-S} & \textbf{Vim-S}  \\
\hline
\textbf{Params(M)} &78.70 & 82.5 & 83.0 & \textbf{29.3} \\ 
\textbf{MU(GB)} &15.83 & 16.54 & 21.80 & \textbf{5.32} \\ 
\textbf{SR} &54.84& 57.75 & \textbf{60.12} & 58.09 \\ 
\textbf{PR} &87.45& 90.69 & \textbf{92.81} & 91.97 \\ 
\toprule
\end{tabular}
\caption{Comparison between different backbone networks on FE240hz dataset. 
`MU' stands for memory usage and `VM' denotes the VMamba model.
} 
\label{tab:CoTMResults} 
\end{table}

\noindent $\bullet$ \textbf{Analysis of Different Backbone Networks.~} 
As illustrated in Table~\ref{tab:CoTMResults}, we conduct the performance analysis of different backbone networks, including Transformer model (ViT-S~\cite{dosovitskiy2020image}, and Swan-S~\cite{liu2021swin}), Mamba backbone networks (VMamba-S~\cite{liu2024vmamba} and Vim-S~\cite{zhu2024vision}), on FE240hz dataset for event camera-based tracking task.
To facilitate information fusion for dynamic template, static template, and search region, We integrate the fusion methods from SwinTrack~\cite{lin2022swintrack} or OSTrack~\cite{ye2022joint} with these backbone networks.
As can be seen from Table~\ref{tab:CoTMResults}, 1) the VM-S model achieves better performance in the SR/PR metric.
2) The Vim-S model obtains competitive performance while maintaining a relatively low parameter count and memory usage.
To strike and balance between computational resource and model performance, we employ Vim-S as the default backbone network in the proposed MambaEVT method.

% compare popular Transformer-based backbones (ViT-B~\cite{dosovitskiy2020image}, ViT-S~\cite{dosovitskiy2020image}, Swin-S~\cite{liu2021swin}) and Mamba-based backbones (VMamba-S, Vim-S) for event tracking on the FE240hz dataset. 
% Since fusion modules are not available for most backbones, we employed the design of SwinTrack\cite{lin2022swintrack} for Swin-Transformer and VMamba, and OSTrack for ViT-S and ViT-B. 

% The results clearly indicate that the parameters, and memory usage of ViT, Swin-Transformer, and VMamba are significantly higher than those of Vim. Although VMamba achieves the highest performance in tracking metrics, it achieves this at the expense of higher resource consumption in parameters and GPU memory compared to Vim.

% In contrast, Vim not only has lower computational and memory requirements but also achieves outstanding performance. This underscores Vim's efficiency and effectiveness as a backbone network, providing an optimal balance between resource consumption and performance.

\noindent $\bullet$ \textbf{Analysis of Memory Library Capacities.~}
To investigate the effect of different long-term (LT) and short-term (ST) memory queue capacities on tracking performance, we conduct analytical experiments on the FE240hz dataset. 
The experimental results are presented in Table~\ref{tab:ablation_memory_queues}.
We can see that the model performance improves with an increase in LT memory queue capacity.
When the LT capacity is set to 16 and the ST capacity to 6, we obtain the best performance, i.e., 58.09 in SR metric and 91.97 in PR metric. 
This finding suggests that enhancing LT capacity allows the network to better capture and utilize long-term dependencies, thereby optimizing tracking performance.

% Table \ref{tab:ablation_memory_queues} explores the impact of varying the capacities of Long-term (LT) and Short-term (ST) memory queues on tracking performance, as measured by SR/PR values on the FE240hz dataset. The results show that the combination of an LT capacity of 16 and an ST capacity of 6 delivers the best performance, achieving a SR/PR of 58.09/91.97.
% The result highlights the importance of integrating a larger LT memory. The Memory Mamba Network exhibits a strong ability to handle longer sequences in the LT module, which significantly enhances performance.
% These findings suggest that increasing the LT capacity enables the network to better capture and leverage long-term dependencies, resulting in improved tracking accuracy and robustness.

\begin{table}
\centering
\small     
\resizebox{\columnwidth}{!}{
\begin{tabular}{cccccccccc}
\toprule 
\textbf{LT/ST}     &6/6 &6/11 &6/16 &11/6 &11/11 &11/16 &16/6 &16/11 &16/16 \\
\midrule
\textbf{SR} &57.83 &57.76 &57.72 &57.39 &57.60 &57.56 &\textbf{58.09} &57.64 &57.75 \\
\textbf{PR} &91.42 &91.65 &91.66 &90.86 &91.26 &91.60 &\textbf{91.97} &91.46 &91.67 \\
\toprule 
\end{tabular}
}
\caption{Ablation study on different capacities of Long-term (LT) and Short-term (ST) memory queues on FE240hz.} 
\label{tab:ablation_memory_queues}
\end{table}

\noindent $\bullet$ \textbf{Analysis of the Number of Dynamic Templates.~} 
As shown in Table~\ref{tab:ablation_template_analysis}, we present the analysis results for different dynamic template number settings on the FE240hz dataset.
We can observe that the performance of the proposed MambaEVT model continues to improve until the number of dynamic templates reaches 11.
When the number of dynamic templates exceeds 11, the model performance declines. 
The possible reason is that the control variable method was applied, meaning the auxiliary parameters were not adjusted to their optimal states corresponding to the increased number of dynamic templates.
Therefore, unless otherwise specified, we set the number of dynamic templates to 11 by default in all experiments.
% 
% the an varying the number of dynamic templates on the FE240hz dataset. It's important to note that the testing phase configurations were kept consistent across all experiments, with the only variable being the number of dynamic templates used during the training phase. The results clearly indicate that using 11 dynamic templates yields the best performance.
% 

\begin{table}[!htp]
    \centering
    \begin{tabular}{c|ccccc}
    \toprule
    \textbf{\# DT}     &5  &7  &9  &11 &13\\
    \hline 
    \textbf{SR}     &0.5627  &0.5665  &0.5602  &\textbf{0.5809} &0.5620\\
    \textbf{PR}     &0.8896  &0.8956  &0.8999  &\textbf{0.9197} &0.8905\\
    \toprule
    \end{tabular} 
    \caption{Analysis of different dynamic templates on FE240hz. `DT' stands for the number of dynamic templates.
    }
    \label{tab:ablation_template_analysis}    
\end{table}

\noindent $\bullet$ \textbf{Analysis of the Number of Layer in Backbone Network.~} 
In this work, we adopt Vision Mamba as the backbone network.
To analyze the impact of the number of Mamba layers on the model parameter efficiency and model performance, we conduct a series of experiments on the FE240hz dataset, as detailed in Table~\ref{tab:ablation_layers_analysis}. 
As the number of layers gradually increases, the performance of our proposed MambaEVT model initially improves before declining.
We obtain the optimal performance when the number of layers is set to 24. 
% This may be because the excessive layers lead to the overfitting of model, which hinders further performance gains.
This is likely because the excessive layers lead to the overfitting of the model, which hinders further performance gains. Based on these findings, we set the layers of Mamba to 24 by default to balance model complexity and tracking effectiveness.

% When performing tracking with Vision Mamba, the number of layers in the Mamba backbone significantly impacts both the parameters and tracking accuracy. In Table \ref{tab:ablation_layers_analysis}, we varied the number of layers to evaluate their effects. During testing, we maintained consistent configurations for the length capacities of both LT and ST. The results indicate that the depth of 24 achieves the best result.
% 

\begin{table} 
    \centering   
    \begin{tabular}{c|cccc}
    \toprule
    \textbf{\#Layers}     &8  &16  &24  &32 \\
    \hline 
    \textbf{Params(M)}  &12.60&20.96 &29.30&37.65 \\
    \textbf{SR}     &0.5687  &0.5746  &\textbf{0.5809}  &0.5756 \\
    \textbf{PR}     &0.8923  &0.9090  &\textbf{0.9197}  &0.9112 \\
    \toprule
    \end{tabular} 
    \caption{Performance metrics for the number of layers in backbone network on FE240hz dataset.}
    \label{tab:ablation_layers_analysis} 
\end{table}

\noindent $\bullet$ \textbf{Analysis of Hyper-parameters.~} 
%\textbf{Analysis of Hyper-parameters.~} 
% 
In this work, we analyze the influence of sampling interval on tracking performance using the EventVOT dataset.
% The sampling interval controls **. 
The sampling interval controls how often the Memory Library strategy uses tracking results. When set to 1, the tracking results are fed into the Memory Library strategy every frame for selection and update. The experimental results are listed in Table~\ref{tab:ablation_sampling_interval}. When the sampling interval are set to 5, the proposed MambaEVT model obtains the optimal performance. Therefore, we set the sampling interval to 5 by default in all experiments.
% The lower bound of ML denotes
% The lower bound of ML denotes the threshold for template similarity used in managing the LT memory library during tracking shifts. If the similarity between a new template and the base template (index 0 in the LT memory library) is below the threshold, the new template is excluded from the LT memory library. Consequently, it does not participate in the Gram matrix calculation.
% To analyze the effect of these hyper-parameters on the final performance, we conduct the ablation study on the EventVOT dataset. The experimental results are listed in Table~\ref{tab:ablation_sampling_interval} and Table~\ref{tab:ablation_thor_lower_bound}. We can find that the model performance is not sensitive to these parameter changes, and the effect is small. When the sampling interval and the lower bound of ML values are set to 5 and 0.35 respectively, the proposed MambaEVT model obtains the optimal performance. Therefore, we set the sampling interval and the lower bound of ML to 5 and 0.35 by default in all experiments.

\begin{table}
    \centering
    \begin{tabular}{c|ccccc}
    \toprule
    \textbf{\#Interval} & 1 & 3 & 5 & 7 \\
    \hline 
    \textbf{SR}     &0.562  &0.563  &\textbf{0.565}  &0.560  & \\
    \textbf{PR}     &0.563  &0.564  &\textbf{0.567} &0.563  & \\
    \toprule
    \end{tabular} 
    \caption{Comparison results of Different Sampling Intervals on EventVOT dataset.}
    \label{tab:ablation_sampling_interval}  
\end{table}

% \begin{table}
%     \centering
%     \begin{tabular}{c|cccccc}
%     \hline 
%     \textbf{LB} & 0.20 & 0.35 & 0.50 & 0.65 & 0.80 & 0.95 \\
%     \hline 
%     \textbf{SR}     & 0.562 & \textbf{0.565} & 0.560 & 0.560 & 0.560 & 0.562 \\
%     \textbf{PR}     & 0.563 & \textbf{0.567} & 0.562 & 0.563 & 0.562 & 0.564\\
%     \hline 
%     \end{tabular} 
%     \caption{Comparison results of Different Lower Bound of ML on EventVOT dataset. `LB' stands for Lower Bound.}
%     \label{tab:ablation_thor_lower_bound}  
% \end{table}

\noindent $\bullet$ \textbf{Analysis on Different Scales of Temporal Windows.~} 
We compare our method using different scales of the temporal window on the EventVOT dataset, as shown in Table~\ref{tab:temporal_window_analysis}. 
Since the optimal timestamp span for frame length is determined empirically and requires further investigation, we scale the original window length to $\times4$ and $\times6$ after extensive trials. 
As we can see from Table~\ref{tab:temporal_window_analysis}, the performance of our proposed MambaEVT model decreases as the temporal window size increases, i.e., from $\times1$ to $\times6$.
These findings indicate that the proposed MambaEVT method is generally effective and that the size of the temporal window significantly affects tracking performance.

\begin{table}[!htp]
    \centering
    \begin{tabular}{c|ccc}
    \toprule
    \textbf{TempW} &$\times1$    &$\times4$  &$\times6$\\
    \hline  
    \textbf{SR}  &0.565    &0.558  &0.557\\
    \textbf{PR}  &0.567    &0.571  &0.569\\
    \textbf{NPR} &0.655     &0.648  &0.645\\
    \toprule
    \end{tabular} 
    \caption{Results of different scales of the Temporal Window (TempW, for short) on the EventVOT.}
    \label{tab:temporal_window_analysis} 
\end{table}

\begin{figure}
    \centering
    \includegraphics[width=\linewidth]{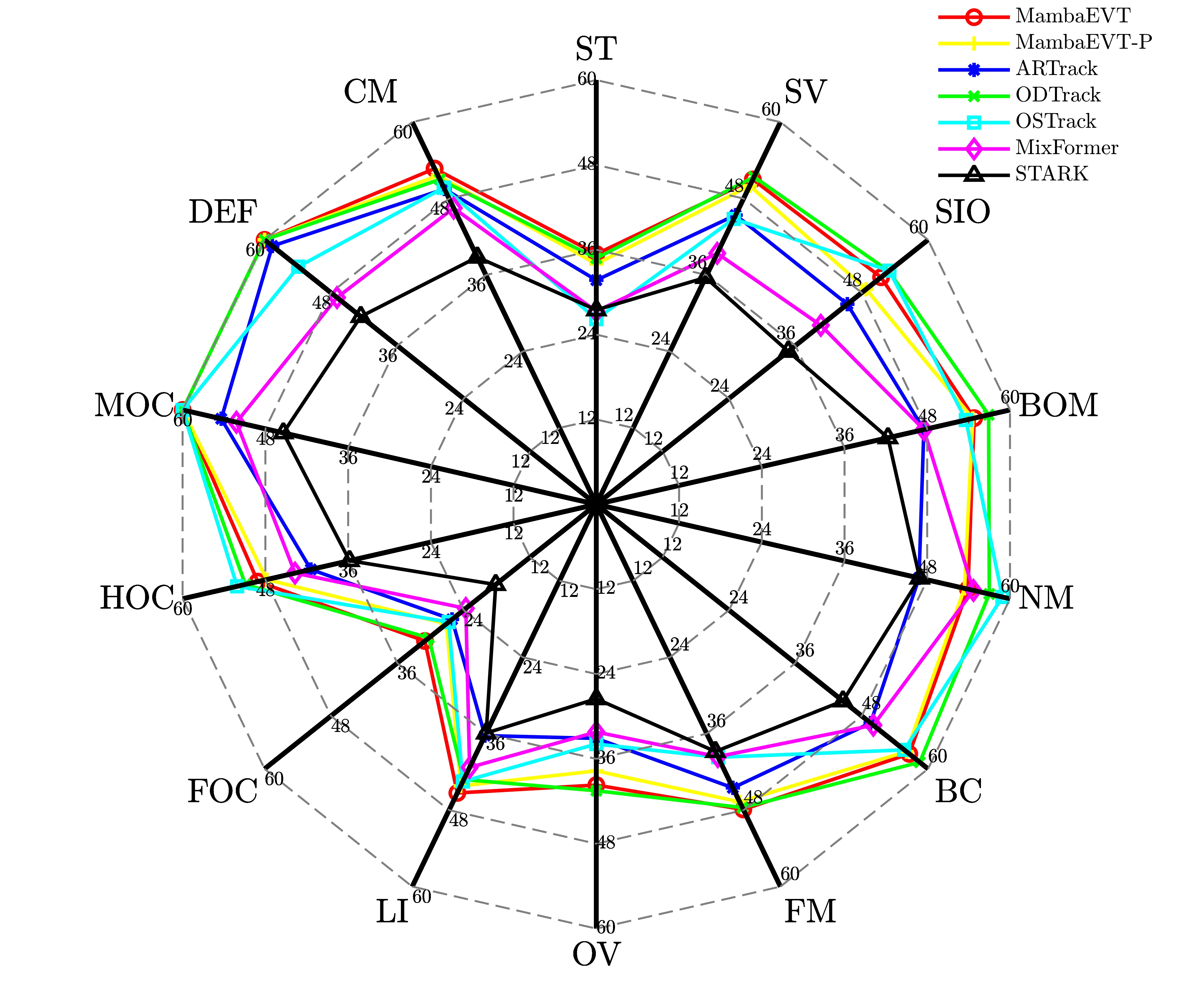}
    \caption{ Tracking results under various challenging factors.}
    \label{fig:EventVOT_radar}
\end{figure}

\noindent $\bullet$ \textbf{Analysis on Tracking in Specific Challenging Environment.~} 
We compare our proposed MambaEVT method with other SOTA trackers across 14 challenging attributes of the EventVOT dataset, such as small target (ST), fast motion (FM), and deformable (DEF), etc. As shown in Figure~\ref{fig:EventVOT_radar}, our proposed MambaEVT model achieves the best performance in some attributes like CM and DEF. Additionally, compared to some other SOTA tracking methods (i.e., ARTrack, ODTrack), we obtain competitive results on some attributes like the influence of background object motion (BOM) and background clutter (BC). These analysis results demonstrate the proposed MambaEVT model offers robust tracking performance across various real-world challenges.

% \subsection{Visualization}

\begin{figure}
    \centering
    \includegraphics[width=\linewidth]{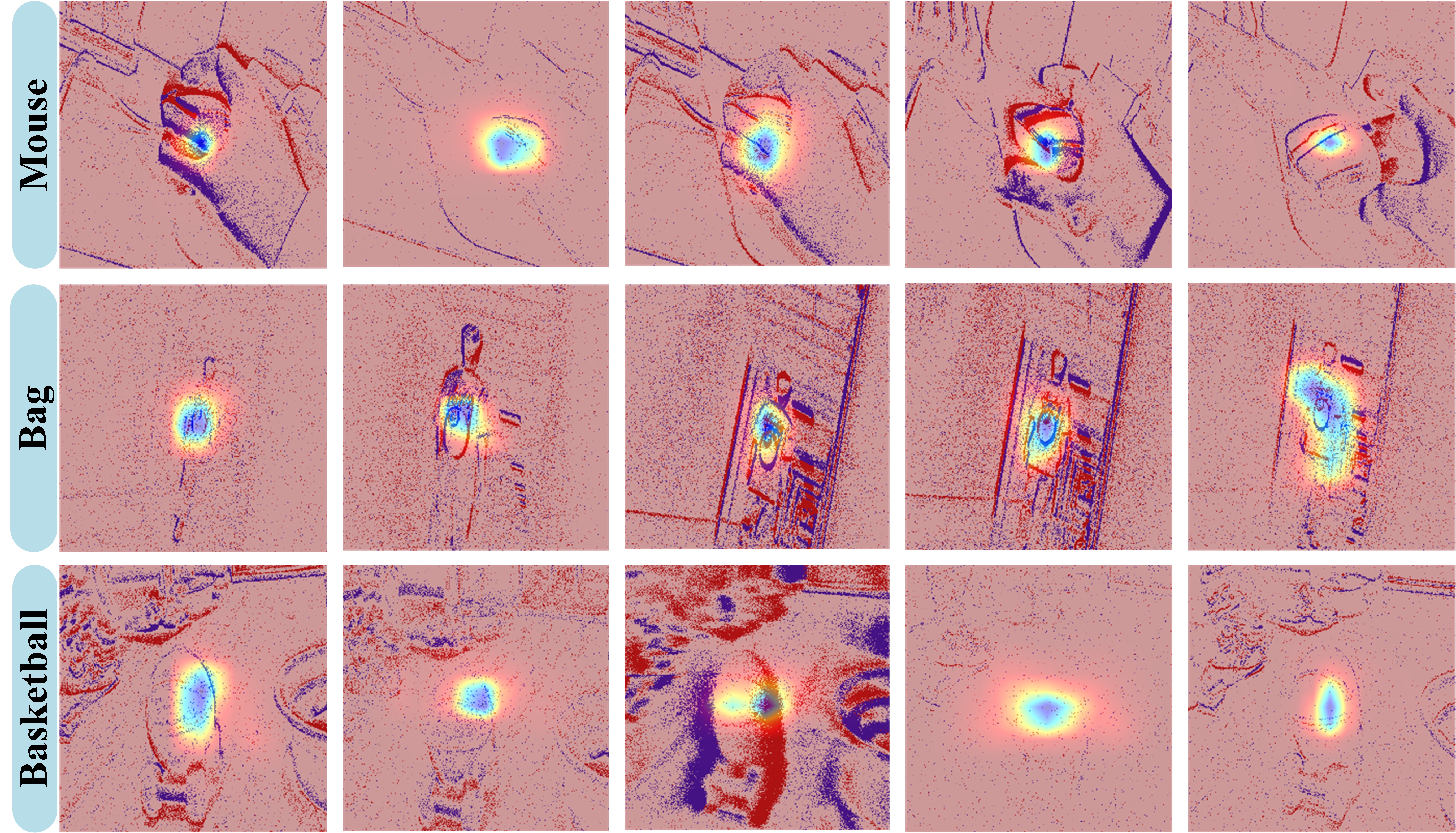}
    \caption{ Activation maps predicted by our proposed MambaEVT framework.}
    \label{fig:activation}
\end{figure}

\begin{figure*}
    \includegraphics[width=\linewidth]{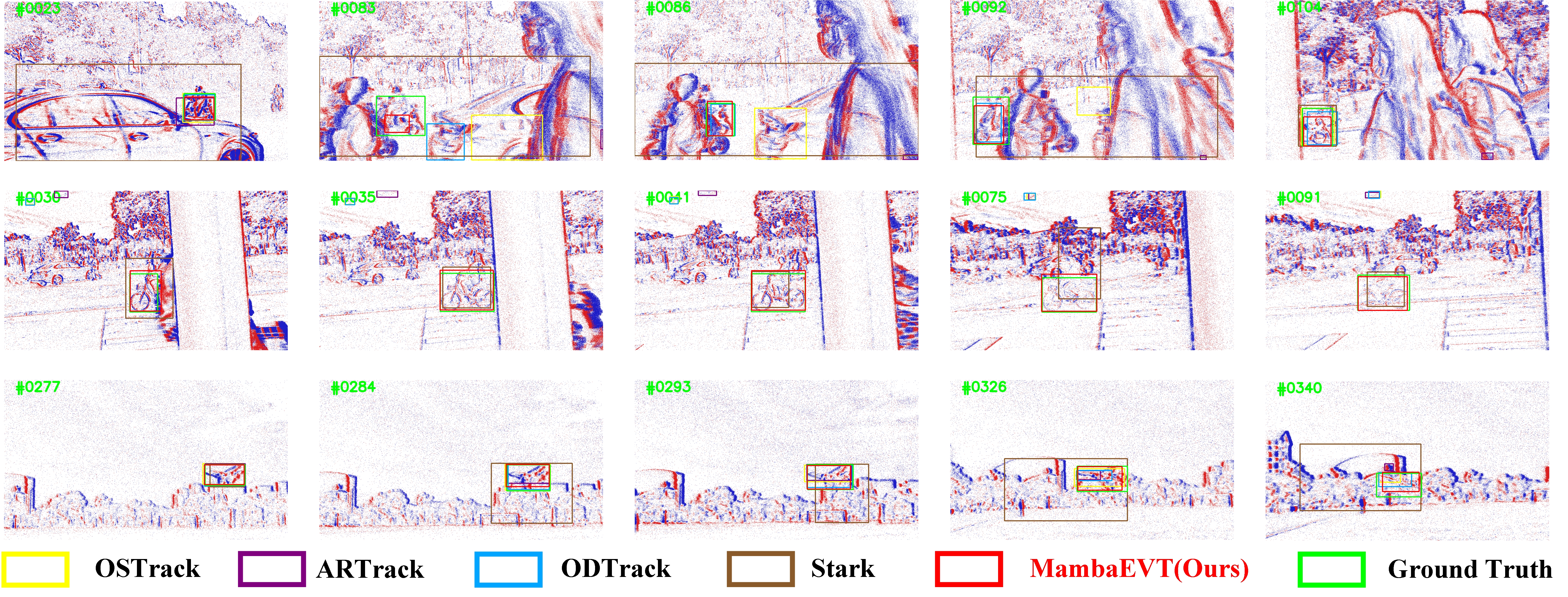}
    \caption{ Visualization of the tracking results of MambaEVT and other SOTA trackers.}
    \label{fig:tracking_result_on_EventVOT}
\end{figure*}

\subsection{Visualization} 

In addition to the numerical evaluations, we carry out a detailed visual examination of the proposed tracking algorithm to facilitate a more profound comprehension of our framework. 
As illustrated in Fig. \ref{fig:tracking_result_on_EventVOT}, we compare the tracking results of our proposed MambaEVT method with other SOTA trackers, such as OSTrack, ARTrack, ODTrack, and Stark, on the EventVOT dataset. 
The results indicate the unique challenges and complexities of tracking with an event camera. 
Although these trackers perform effectively in simple scenarios, there remains a substantial opportunity for enhancement. 
Additionally, we visualize the response maps of the tracking head on multiple videos, including Mouse, Bag, and Basketball.  
As illustrated in Fig.~\ref{fig:activation}, our tracker effectively highlights the target object regions, confirming its ability to accurately focus on the targets. This effectiveness is particularly evident in scenarios with significant background changes, where our tracker maintains strong performance by consistently distinguishing the target from its surroundings.

\subsection{Limitation Analysis}  
Although our tracker achieves a good balance between performance and the number of parameters, it suffers from a relatively low FPS. 
Further design improvements are necessary to enhance the tracking speed. 
Additionally, as the Mamba model is designed for long sequence modeling, our current training process may not fully leverage its capabilities, potentially limiting its overall performance.

\section{Conclusion}  
The Mamba-based visual tracking framework introduced in this paper represents a significant advancement in event camera-based tracking systems. We have achieved simultaneous feature extraction and interaction for search regions and target templates by leveraging the state space model with linear complexity and lead to improved target localization. The integration of a dynamic template update strategy through the Memory Mamba network has proven to be a crucial step in enhancing tracking performance. By dynamically adjusting the template memory module to account for the diversity of samples in the target template library, we have been able to maintain a more effective dynamic template. This innovative approach has allowed our tracking algorithm to strike a favorable balance between tracking accuracy and computational efficiency. The robustness and effectiveness of our method have been demonstrated across multiple large-scale datasets, including EventVOT, VisEvent, and FE240hz. We believe that this work not only contributes to the current state-of-the-art in event-based visual tracking but also paves the way for future research in this field. The source code of this work will be released to facilitate further research and development in event camera-based tracking systems. In our future work, we will further improve the running efficiency of our proposed tracker and decrease the energy consumption in the tracking phase (by spiking neural networks).

% \section*{Acknowledgement}   

{
\small
\bibliographystyle{ieeenat_fullname}
\bibliography{reference}
}

\end{document}